\documentclass[journal]{IEEEtran}

\usepackage[english]{babel}
\usepackage[utf8]{inputenc}
\usepackage{cite}

\usepackage[pdftex]{graphicx}

\usepackage[cmex10]{amsmath}
\usepackage{amssymb}

\usepackage{ctable}

\usepackage{array}

\ifCLASSOPTIONcompsoc
  \usepackage[caption=false,font=normalsize,labelfont=sf,textfont=sf]{subfig}
\else
  \usepackage[caption=false,font=footnotesize]{subfig}
\fi

\usepackage{dblfloatfix}

 \let\MYorigsubfloat\subfloat
 \renewcommand{\subfloat}[2][\relax]{\MYorigsubfloat[]{#2}}

\usepackage{hyperref}

\usepackage{booktabs}

\newcommand{\I}{I}

\newcommand{\beq}{\begin{equation*}}
\newcommand{\eeq}{\end{equation*}}
\newcommand{\beqn}{\begin{equation}}
\newcommand{\eeqn}{\end{equation}}

\begin{document}
%
\title{Data-Driven Color Augmentation Techniques \\ for Deep Skin Image Analysis}
%
%
%

\author{Adrian~Galdran$^{*,1}$, Aitor Alvarez-Gila$^{2,3}$, Maria Ines Meyer$^{1}$, Cristina L. Saratxaga$^2$, Teresa Ara\'ujo$^{1,4}$, \\
 Estibaliz Garrote$^2$, Guilherme Aresta$^{1,4}$, Pedro Costa$^1$, A.M. Mendon\c{ç}a$^{1,4}$, and~Aur\'elio Campilho$^{1,4}$
\thanks{A. Galdran$^*$, M. I. Meyer, T. Ara\'ujo, G. Aresta, P. Costa, A.M. Mendon\c{ç}a, and A. Campilho are with INESC TEC Porto, Portugal; e-mails: \{adrian.galdran, maria.i.meyer, tfaraujo, guilherme.m.aresta, pvcosta\}@inesctec.pt.}
\thanks{A. Alvarez-Gila, C. L. Saratxaga, and E. Garrote are with Tecnalia, Spain; e-mails: \{aitor.alvarez, cristina.lopez, estibaliz.garrote\}@tecnalia.com}
\thanks{A. Alvarez-Gila is also with the Computer Vision Center, Universitat Aut\'onoma de Barcelona, Spain.}
\thanks{A,M. Mendon\c{ç}a and A. Campilho are also with Faculdade de Engenharia, Universidade do Porto, Portugal; e-mails: \{amendon, campilho\}@fe.up.pt.}}

%

\maketitle

\begin{abstract}
Dermoscopic skin images are often obtained with different imaging devices, under varying acquisition conditions. In this work, instead of attempting to perform intensity and color normalization, we propose to leverage computational color constancy techniques to build an artificial data augmentation technique suitable for this kind of images. Specifically, we apply the \emph{shades of gray} color constancy technique to color-normalize the entire training set of images, while retaining the estimated illuminants. We then draw one sample from the distribution of training set illuminants and apply it on the normalized image. We employ this technique for training two deep convolutional neural networks for the tasks of skin lesion segmentation and skin lesion classification, in the context of the ISIC 2017 challenge and without using any external dermatologic image set. Our results on the validation set are promising, and will be supplemented with extended results on the hidden test set when available.
\end{abstract}


\IEEEpeerreviewmaketitle

\section{Introduction}
\IEEEPARstart{M}{elanoma} is a highly aggressive form of skin tumor that can be successfully treated when diagnosed at an early stage \cite{meckbach_survival_2014}. Suspicious symptoms can be detected visually, and for that reason computer-vision techniques applied to skin image analysis are becoming increasingly important as a feasible and inexpensive method for skin care \cite{celebi_dermoscopy_2015}.

For an effective interpretation of skin lesions, color is known to be a fundamental visual feature, together with size, shape and texture \cite{abbasi_early_2004}. 
In computerized skin image analysis, a representation of the image in terms of visual features is usually built in order to find relevant cues for diagnosis. Deep learning techniques, which can automate the feature design step, are outperforming traditional, hand-crafted features-based methods in most computer-aided skin disease diagnosis tasks \cite{codella_deep_2016,yu_automated_2016}. However, training these techniques requires large databases of labeled skin images. These databases often contain samples obtained at different hospitals, with different acquisition systems and under varying illumination conditions, which poses a complex challenge for robust image analysis. 

Usually, the input of skin image understanding techniques undergoes a color normalization step, in order to achieve color constancy \cite{gijsenij_computational_2011}. This has been recently shown to be helpful for subsequent automatic diagnosis tasks \cite{barata_improving_2015}. This procedure is related to the need of handcrafted feature-based methods to receive normalized input images. This is especially important at test time, since feeding the model images that are significantly different from those that the algorithms saw during training time can lead to a poor generalization of the method, due to their limited expressiveness. However, computational color constancy is a hard problem, and the output can sometimes be unpredictable. 


In this work, we propose a data augmentation technique adapted for skin lesion analysis with deep neural networks, leveraging the fact that their virtually unlimited expressive power enables them to be trained with extensive variability in the data manifold, in order for them to learn features invariant to those factors of variation. Specifically, instead of relying on existing illuminant color compensation techniques to achieve a normalized dataset, we apply them to obtain an estimate of the color of the illuminant of the scene. These illuminants are then sampled at training time, and applied to white-balanced training images in order to generate new training images that simulate different but plausible illumination conditions at acquisition time. This augmentation technique leads to a robust training stage, which allows for promising results to be obtained both in segmentation and classification tasks. Our work is framed into the ISIC 2017 Challenge: \emph{Skin Lesion Analysis-Towards Melanoma Detection} \cite{_international_????}. The competition provides three different goals, from which we only consider skin lesion segmentation and classification. 

\section{Data-Driven Artificial Data Augmentation}\label{sec:data_augmentation}
In order to train complex models with a small amount of data, image augmentation schemes are often applied. 
These consist of transforming input images into new but plausible examples that can help the network to generalize better. Although recent generative models suggest that realistic medical image generation is feasible by means of more complex techniques \cite{costa_towards_2017}, typically basic linear transformations on the image geometry are performed, such as shifting, scaling or rotating. 
However, applying domain-specific knowledge on the expected images, \emph{e.g.} on the image acquisition process, can help to generate a more diverse set of artificial images while keeping plausibility. 

Below we describe a method to jointly normalize color in skin images and learn from the available data the distribution of scene illuminants.
These illuminants are then applied to white-balanced images for generating useful artificial images.

\begin{figure}[t]
\begin{center}
\includegraphics[width =0.45 \textwidth]{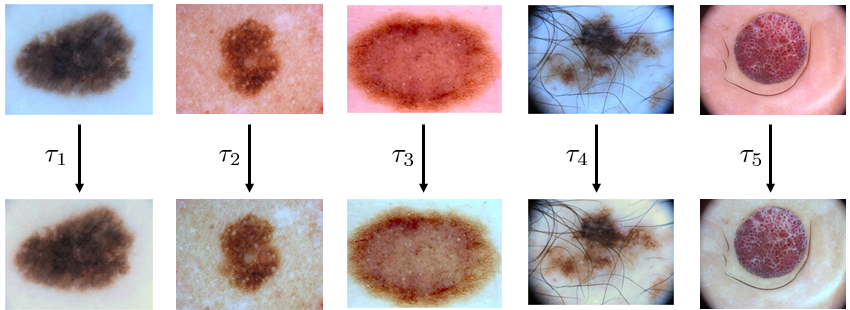}
\end{center}
\caption{White-balancing and illuminant estimation on several training images following Eq. \ref{white_bal}}
\label{fig_skin_application}
\end{figure}


\subsection{Color Image Augmentation via Illuminant Estimation}
A well-known property of the human visual system is the ability to perceive the color of an object as roughly constant, even when the color of the light projected on the scene is modified. 
This effect is known as \emph{color constancy} and, although not yet fully understood, it can be partially explained by the joint interaction of low level mechanisms that act on the response of the cones in the retina and by higher cognitive-level effects \cite{fairchild_color_2013}. 
In parallel, there exists a large number of techniques that attempt to reproduce such effect on image acquisition and processing systems within the field of computational color constancy. 
In this context, the goal is often to first estimate the illuminant of the scene, and then remove it from the input image by means of a chromatic adaptation transform. 
Usually this is achieved in commercial cameras by means of proprietary white-balancing algorithms. 
However, this general approach does not always work in a predictable manner, as the illuminant depends not only on the light cast on the scene but also on the image content.

\begin{figure}[t]
\begin{center}
\includegraphics[width = 0.85\columnwidth]{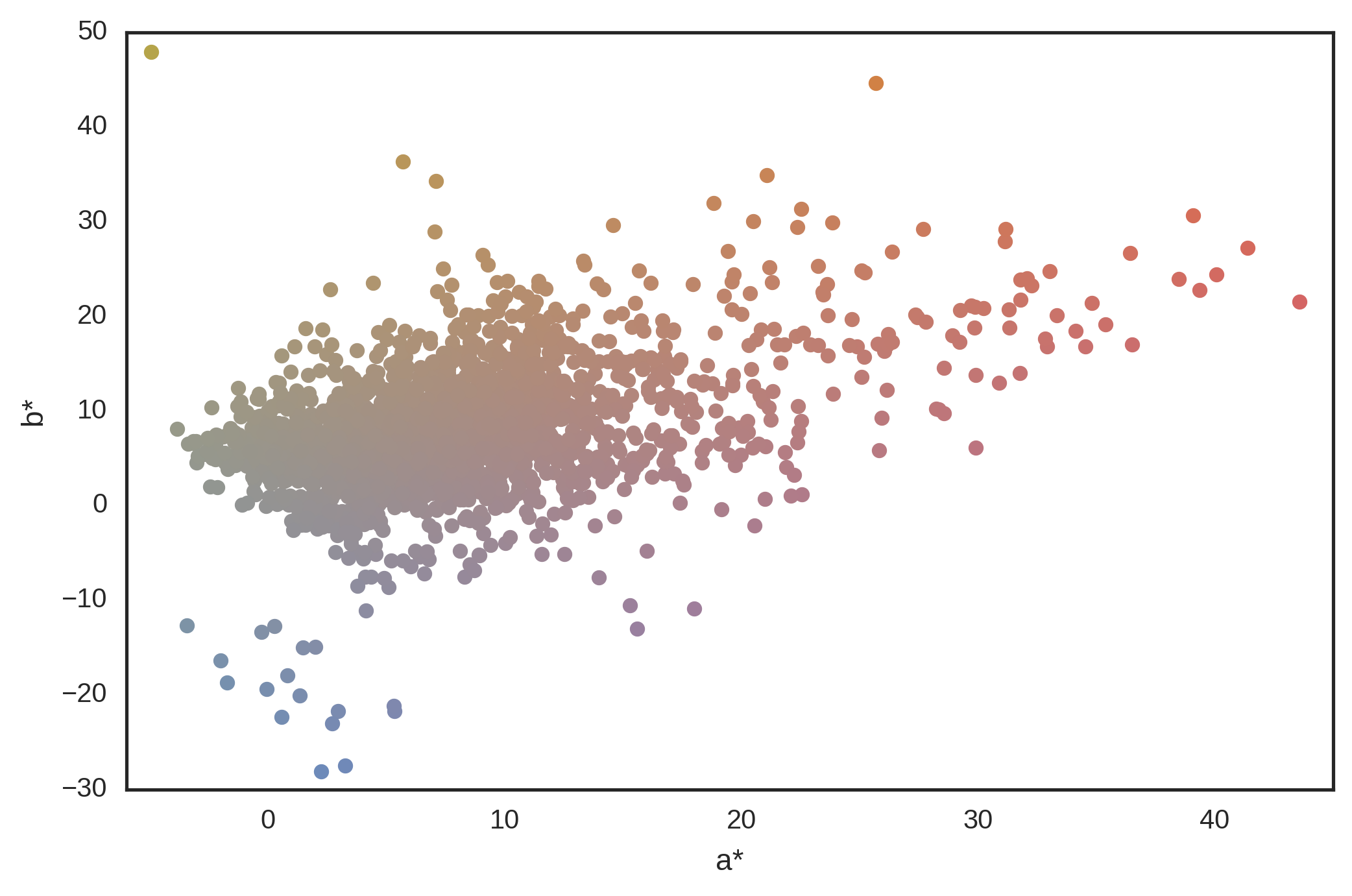}
\end{center}
\caption{Distribution of illuminants estimated from the training set using the \emph{shades of gray} algoritm. The RGB illuminants where converted to the CIE L*a*b* uniform color space \cite{robertson_cie_1977} via their XYZ tristimulus values, and projected onto the a*b* plane (a* approximates redness-greenness, b* approximates yellowness-blueness). Each sample shows the color it encodes.}
\label{fig_train_set_illum_distrib}
\end{figure}

A white-balancing technique mainly works by manipulating image intensities in such a way that the scene appears as being lit by a neutral illuminant.
This is usually accomplished separately in the different spectral components of the image. 
Denoting an image as $I=(I^R, I^G, I^B)$, a typical white balancing technique solves the following equation:
\begin{equation}\label{white_bal}
I^{\lambda}_j(x) = \tau^\lambda \cdot I^\lambda_{wb}(x),
\end{equation}
where $\lambda\in\{R,G,B\}$ and $I_{wb}(x)$ is the white-balanced version of $I(x)$, which can be recovered by inversion of Eq. (\ref{white_bal}). 
In this work, we estimate the illuminant on images from the training set applying the \textit{Shades of Gray} color constancy technique \cite{finlayson_shades_2004}, as implemented by the general color constancy framework described in \cite{van_de_weijer_edge-based_2007}. 
A sample of training images and their white-balanced versions is shown in Fig. \ref{fig_skin_application}, while Fig. \ref{fig_train_set_illum_distrib} shows the full empirical distribution over the illuminants. 

\begin{figure}[t]
\begin{center}
\includegraphics[width =0.4 \textwidth]{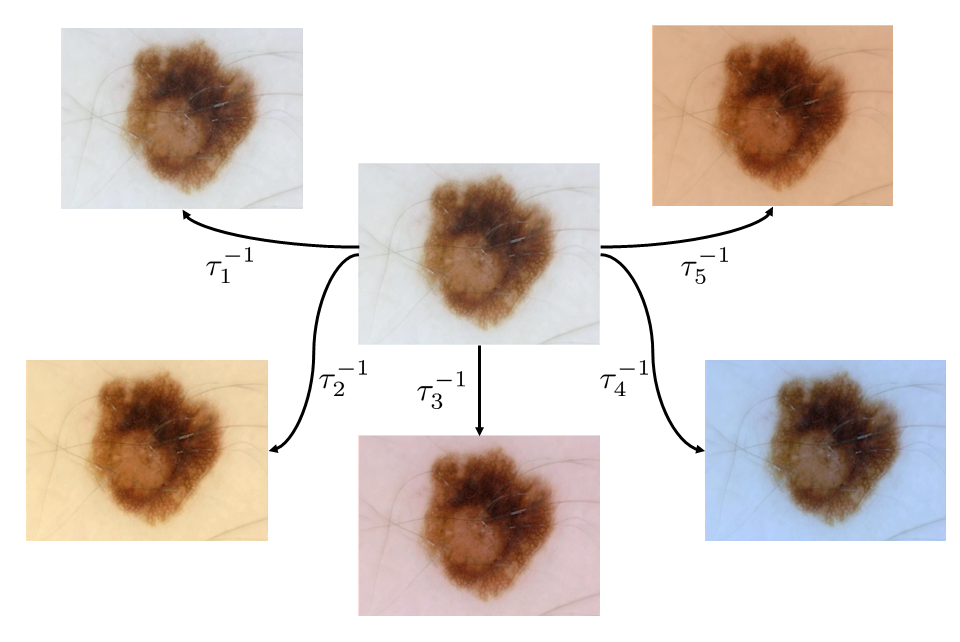}
\end{center}
\caption{Color-casting of a white-balanced training image with different illuminants.}
\label{fig_skin_application2}
\end{figure}

Note that in \cite{barata_improving_2015} the same approach was successfully employed to normalize skin images before supplying them to a classifier, observing an improvement in performance after this normalization. 
In this work, we proceed in a substantially different direction. 
Instead of only removing illuminant variations, we collect the set of illuminants extracted from the training set, $\mathcal{T} = \{\tau_i^{-1}, \ i = 1,\ldots, 2000\}$. 
At training time, we augment the white-balanced dataset by color-casting each corrected image with an illuminant extracted from $\mathcal{T}$, i.e., $I_{new}(x) = \tau^{-1} \cdot \I_{wb}(x)$. This produces realistically rendered versions of each of them under a different illuminant. 
This technique is applied in an online fashion (randomly yielding one version of each image per epoch) and by applying a Von Kries-like diagonal chromatic adaptation transform \cite{von1970influence}. 
Notice that, if the illuminant $\tau$ is the same that was extracted from $I$, this transformation reduces to the identity, \emph{i.e.}, $I_{new}(x) = I(x)$. 
However, if $\tau$ comes from a different image, this transformation leads to a color-casted version of $I$. 
The result of applying this procedure to a white-balanced image from the training set can be observed in Fig. \ref{fig_skin_application2}. 
With this strategy, the method is able to realistically augment the original dataset, preventing overfitting when training a deep neural network and achieving a more robust training procedure. 

After extracting the set of illuminants from the training data, we need to access its underlying probabilistic distribution in order to generate new plausible images. 
The way in which that distribution is modeled is relevant to the quality of the obtained artificial images. 
In \cite{lou_color_2015}, a similar technique was applied to augment the dataset before learning a deep model that can achieve color constancy on natural images. 
In that work, the authors applied a $k$-means clustering to the retrieved illuminants, with $k$ being an heuristically determined free parameter. 
However, this approach may lead to an over-representation of areas in the illuminant space that, although representing a separate cluster, do not contain a significant amount of samples. 
Here we apply a different strategy and directly sample from the raw empirical distribution of illuminants. 
This way, each time a training example is selected we randomly choose an illuminant from $\mathcal{T}$ with a uniform probability distribution, producing a new color-casted image. 

\subsection{Gamma-Correction Augmentation}
In addition to the white-balancing, a digital camera applies several other normalization steps \cite{bertalmio_image_2014}. 
One of them is the well-known gamma correction. 
When a camera captures light input, theoretically the received signal is a linear function of photons reaching the device's sensors. 
However, for a more natural luminance reproduction in digital screens, a non-linear transformation with a power function is usually applied:
\begin{equation}\label{gamma}
I(x) = L(x)^\gamma,
\end{equation}
where $L(x)$ is the luminance reaching the camera, $I(x)$ is the corresponding post-processed image ready for display, and $\gamma$ is the correction constant.
Unfortunately, the specifications of this transform depend heavily on the characteristics of the display and on the imaging device manufacturer. 
To compensate for this issue, we also applied gamma corrections to the training images both at train and test time, aiming for a more robust representation learning of our model. 
Thus, after undergoing the color transformation explained above, we randomly draw for each image and epoch a correction factor $\gamma$ from a Normal distribution of mean $\mu=1$ and standard deviation $\sigma = 0.1$ truncated at 0 and 2, and apply a power-law mapping similar to the one in Eq. (\ref{gamma}).

\subsection{Non-Linear Geometrical Image Augmentation}
To complement the color transformation detailed above, we applied standard geometrical (linear) data augmentation techniques, namely rotation, horizontal and vertical flipping, translation and scaling of the input image. 
Moreover, to account for the non-linear distortions typical of soft tissues such as skin, we applied several non-linear transformations of the image geometry, similar to those proposed in \cite{codella_deep_2016}.

\section{Deep Neural Networks for Skin Lesion Segmentation}
\begin{figure}[t]
	\begin{center}
		\includegraphics[width =0.49 \textwidth]{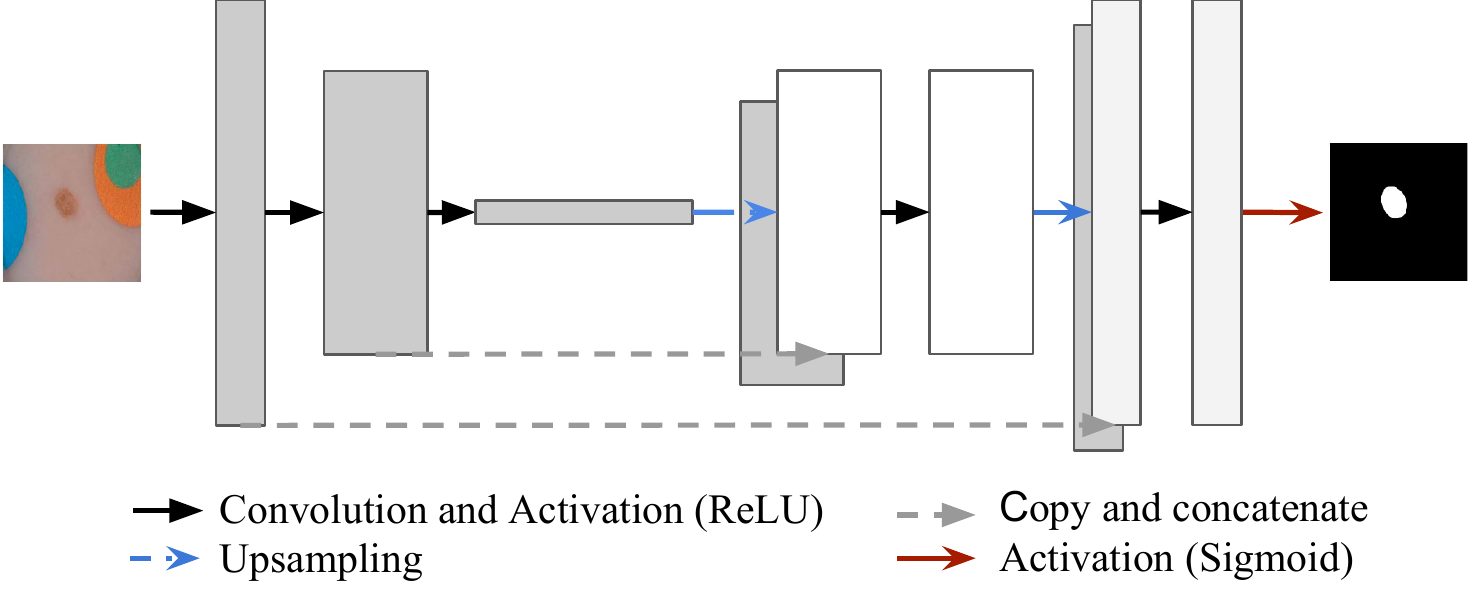}
	\end{center}
	\caption{Schematic representation of the U-Net architecture.}
	\label{fig_unet_architecture}
\end{figure}

The described lesion segmentation problem is approached using a Convolutional Neural Network (CNN), coupled with extensive data augmentation. Data was augmented applying the transforms described in Section \ref{sec:data_augmentation} and re-scaled before being employed as input to a U-Net. The U-Net architecture was first proposed in \cite{ronneberger_u-net:_2015}, and is a Fully Convolutional Network, originally designed for segmenting neuronal structures and cells in microscopy images. It is a powerful deep classifier, that can be successfully trained with a relatively low amount of training data and still produces accurate segmentations. 

The architecture of the U-Net is represented in Fig. \ref{fig_unet_architecture}. The network has two main paths: a contracting path, and a dimensionally symmetric expanding path.
The contracting path consists of consecutive convolutional layers with a stride of $2$. Each convolutional layer is followed by a Rectified Linear Unit (ReLU) activation and batch normalization, except for the last layer, which is activated by a sigmoid activation function. 
The stride of the convolutional layers is selected so that the dimension of the output feature map of the contracting path layers decreases until $1 \times 1 \times nf $, where $nf$ is the number of filters.
This point in the network marks the beginning of the expanding path. The output of each layer is upsampled so that it has the same dimension as the corresponding layer in the contracting path. To compensate for the loss in spacial resolution that results from the multiple downsampling operations, the upsampled feature map is concatenated with the feature map of the corresponding layer in the contracting path. 

The network was trained using gradient descent backpropagation, with the Adam optimizer and a Jaccard index-based loss function, as in \cite{kayalibay_cnn-based_2017}. 
The system outputs the probability of each pixel of the input image belonging to a lesion.

\section{Deep Neural Networks for Skin Lesion Classification}

The image classification task in the ISIC 2017 Challenge comprises two independently evaluated binary classification subtasks over the same set of images: 1) discriminating melanoma vs. all the other lesions and 2) discriminating seborrheic keratosis from every other kind of deseases.  
Both tasks were independently approached by means of separately trained networks. 
Nevertheless, as a common strategy, and following the rationale in \cite{yu_automated_2016}, we chose to leverage the outcome from the lesion segmentation task by feeding the classification stage with images cropped around the bounding box defined by the segmentation mask. 
At train time we used the ground truth masks and increased the tight bounding box by $15\%$ for incorporating information on the appearance of the skin around the lesion, which may contain useful information for diagnosis. 
At test time we relied on the predicted segmentations and applied the same margin, which can also help alleviate small segmentation errors and keep lesion border information. 

We trained both tasks with $50$ layer versions of deep residual networks \cite{he_deep_2016}. These were initialized with the weights learned by pre-training over the ILSVRC2012 Imagenet database. We replaced the last fully connected layer of the pretrained network by a dropout stage and two dense layers of $1024$ and $2$ outputs, respectively, followed by the final softmax. 
Given the strong tendency to overfit, a dropout probability of $0.8$ was set, and a two-phase fine-tuning was performed. During the first stage, only the new dense layers were trained (using an Adadelta optimizer). During the second phase, the first $47$ layers were still kept frozen, and the rest were trained using stochastic gradient descent with a learning rate of $0.001$ and momentum ($0.9$). The loss function was categorical cross entropy, but the losses associated with samples from each class were weighted differently, to compensate for class imbalance.

\section{Experimental Evaluation}
Both models were tested on the ISIC 2017 Challenge. The organizers provided a training set of $2000$ images with ground-truth for segmentation and classification, a validation set of $150$ images, and a test set with $500$ images. It is important to stress that we did not employ any external skin image set or categorical data (other than natural images from the Imagenet dataset in the Resnet50 pretraining). This was allowed by the challenge organizers, but we wanted to explore the capability of the proposed technique to improve results when only few domain-specific images are available for training.

For performing predictions on unseen images, we applied randomly selected illuminants to the white-balanced input image, and predicted on several color casted versions of it. The overall prediction for each pixel was computed as the median of the predictions on each tested image. No further post-processing was performed in the output.

Table~\ref{tab_results_segmentation} shows the results obtained for the segmentation task on the validation set, while Table~\ref{tab_results_classification} does so for the classification task.
Results on the test set will be supplied for both tasks as soon as they are available.

\begin{table}[t]
  \caption{\hspace{1cm} Results for the segmentation task \newline \scriptsize{ Acc = Accuracy, DC = Dice Coefficient, JD = Jaccard Distance \newline Ss = Sensitivity, Sp = Specificity}}
  \label{tab_results_segmentation}
  \centering
  \begin{tabular}{llllll}
      \toprule
      Task & ACC & DC & JD & SS & SP \\
      \midrule
      1) Segmentation & 0.948 & 0.846 & 0.767 & 0.865 & 0.980 \\
      \bottomrule
    \end{tabular}
\end{table}

\begin{table}[t]
  \caption{\hspace{1cm} Results for the classification task \newline \scriptsize{AUC = Area Under the ROC curve, AP = Average Precision \newline ACC, AP, SS and SP were computed at 0.5 confidence threshold}}
  \label{tab_results_classification}
  \centering
  \begin{tabular}{llllll}
    \toprule
    Task & AUC & ACC & AP & SS & SP \\
    \midrule
    3.1) Melanoma & 0.791 & 0.580 & 0.482 & 0.833 & 0.517 \\
    3.2) Seborrheic keratosis & 0.954 & 0.867 & 0.898 & 0.857 & 0.870 \\
      \midrule
    Average & 0.873 & 0.723 & 0.690 & 0.845 & 0.694 \\
    \bottomrule
  \end{tabular}
\end{table}

\subsection{Discussion and Conclusion}
In this paper we presented our approach to the Segmentation and classification tasks of the ISIC 2017 Challenge \emph{Skin Lesion Analysis - towards melanoma detection}. We focused our efforts on exploiting the information obtained from applying color constancy algorithms to the training set in order to perform extensive data augmentation that could help regularize our convolutional neural networks-based learning pipeline. The obtained segmentations were leveraged to improve the full image classification, obtaining competitive results, especially on the segmentation task.

The strong shift of the distribution in Fig. \ref{fig_train_set_illum_distrib} towards the $a*>0, b*>0$ quadrant suggests that part of the illuminant estimation may be inaccurately considering the skin's color as being produced by an unnatural reddish illuminant in some cases. In the future, we aim at further investigating this effect.

\bibliographystyle{IEEEtran}
\bibliography{skin_image_analysis_refs}

\end{document}